\documentclass[conference]{IEEEtran}
\usepackage{cite}

\ifCLASSINFOpdf
   \usepackage[pdftex]{graphicx}
   \graphicspath{{Medias/}}
  \DeclareGraphicsExtensions{.pdf,.jpeg,.png}
\else
  \usepackage[dvips]{graphicx}
  \DeclareGraphicsExtensions{.eps}
\fi
\usepackage{stmaryrd} 
\usepackage{dsfont} 

\usepackage[cmex10]{amsmath}
\usepackage[standard]{ntheorem}
\begin{document}
%
\title{Building a Chaotic Proved Neural Network}

\author{\IEEEauthorblockN{Jacques M. Bahi, Christophe Guyeux, and Michel Salomon}
\IEEEauthorblockA{Computer Science Laboratory (LIFC)\\
University of Franche-Comt\'e\\
IUT de Belfort-Montb\'eliard \\
BP 527, 90016 Belfort Cedex, France\\
Email: christophe.guyeux@univ-fcomte.fr}
}


%


\maketitle

\begin{abstract}

Chaotic neural networks have received  a great deal of attention these
last  years.  In  this  paper we  establish  a precise  correspondence
between  the so-called chaotic  iterations and  a particular  class of
artificial neural networks:  global recurrent multi-layer perceptrons.
We show formally  that it is possible to  make these iterations behave
chaotically,  as defined  by Devaney,  and  thus we  obtain the  first
neural networks proven chaotic. Several neural networks with different
architectures are trained to exhibit a chaotical behavior.

\end{abstract}


%
\IEEEpeerreviewmaketitle

\section{Introduction}

Due to the widespread use of the Internet and new digital technologies
in nowadays life, security in computer applications and networks never
was such a hot  topic.  Digital rights managements, e-voting security,
anonymity  protection, and  denial  of services  are  examples of  new
security  concerns appeared  this last  decade.  Tools  on  which this
security  is based  are, among  others: hash  functions, pseudo-random
number    generators,   cryptosystems,   and    digital   watermarking
schemes. Due to their wide  use in security protocols, these tools are
targeted everyday by hackers  and new threats are frequently revealed.
For  example, security  flaws  have been  recently  identified in  the
previous standard in hash  functions called SHA-1 \cite{rijmen05}.  As
the  new standards  (SHA-2  variants) are  algorithmically similar  to
SHA-1, stronger hash functions using new concepts are desired.

New  approaches   based  on  chaos  are  frequently   proposed  as  an
alternative to solve concerns which recurrently appear in the computer
science security field \cite{guyeux10,bg10:ip,wbg10:ip}. The advantage
of the  use of  chaotic dynamics for  security problems lies  in their
unpredictability proved by the mathematical theory of chaos.
This  theory brings  many qualitative  and quantitative  tools, namely
ergodicity, entropy,  expansivity, and sensitive  dependence to initial
conditions \cite{GuyeuxThese10}.  These tools  allow the study  of the
randomness  of  the  disorder   generated  by  the  considered  system
\cite{guyeuxTaiwan10}.

Recently, many researchers have built chaotic neural networks in order
to   use  it   as  a   component  of   new  proposed   hash  functions
\cite{springerlink:10.1007/s00521-010-0432-2},   pseudo-random  number
generators,  cryptosystems \cite{dalkiran10,Lian20091296},  and digital
watermarking schemes.   Since the first introduction  by McCulloch and
Pitts  in 1959,  artificial  neural  networks have  been  shown to  be
efficient  non-linear  statistical   data  modeling  tools  which  can
implement complex  mapping functions.  Hence,  they may be  trained to
learn a  chaotic process and  also, by construction,  exhibit suitable
properties:  data confusion  and diffusion,  one-way function and compression.
Security is  not the  only application domain  of such new  tools: the
existence of  chaos in our brain  has been recently  revealed, and the
use of a  chaotic artificial neural network as a  model can serve, for
example, neuroscientists in their attempts to understand how the brain
works.

However, using an element of chaos as a component of the scheme is not
sufficient, in our opinion, to be able to claim that the whole process
behaves chaotically.  We believe that this claim is not so evident and
must be  proven. Let us notice  that up to now  the proposed chaotical
neural networks have failed  to convince the mathematics community due
to a lack of  proof. This is why it is explained  in this paper how it
is  possible  to  build  an  artificial neural  network  that  behaves
chaotically,  as it  is defined  by Devaney  \cite{Devaney}.   We will
establish  a  correspondence between  particular  neural networks  and
chaotic  iterations,  which  leads  to  the definition  of  the  first
artificial neural network proven chaotic, according to~Devaney.

The remainder of this paper  is organized as follows. The next section
is devoted to some recalls  on chaotic iterations and Devaney's chaos,
followed by a brief  description of artificial neural networks (ANNs).
Section~\ref{sec:RelatedW} presents a review  of some works related to
chaotic neural  networks.  Our approach, which consists  in building a
global recurrent  ANN whose iterations are chaotic,  is formalized and
discussed  in Section~\ref{sec:KoANN}.  Concrete  examples of  chaotic
neural networks  also show  the relevance of  our method.   Finally in
Section~\ref{sec:Conclusion} we conclude and outline future work.

\section{Basic Recalls}

In the sequel $S^{n}$ denotes the  $n^{th}$ term of a sequence $S$ and
$E_{i}$ denotes the $i^{th}$  component of a vector $E$. $f^{k}=f\circ
...\circ   f$  is for   the  $k^{th}$   composition  of   a  function
$f$.     Finally,     the      following     notation     is     used:
$\llbracket1;N\rrbracket=\{1,2,\hdots,N\}$.

\subsection{Chaotic iterations versus Devaney's chaos}

\subsubsection{Chaotic Iterations}
\label{sec:chaotic iterations}

Let us consider  a \emph{system} with a finite  number $\mathsf{N} \in
\mathds{N}^*$ of elements  (or \emph{cells}), so that each  cell has a
boolean  \emph{state}. A  sequence of  length $\mathsf{N}$  of boolean
states of  the cells  corresponds to a  particular \emph{state  of the
system}. A sequence which  elements belong to $\llbracket 1;\mathsf{N}
\rrbracket $ is called a \emph{strategy}. The set of all strategies is
denoted by $\mathbb{S}.$

\begin{definition}
\label{Def:chaotic iterations}
The      set       $\mathds{B}$      denoting      $\{0,1\}$,      let
$f:\mathds{B}^{\mathsf{N}}\longrightarrow  \mathds{B}^{\mathsf{N}}$ be
a  function  and  $S\in  \mathbb{S}$  be  a  strategy.  The  so-called
\emph{chaotic      iterations}     are     defined      by     $x^0\in
\mathds{B}^{\mathsf{N}}$ and
$$     
\forall    n\in     \mathds{N}^{\ast     },    \forall     i\in
\llbracket1;\mathsf{N}\rrbracket ,x_i^n=\left\{
\begin{array}{ll}
  x_i^{n-1} &  \text{ if  }S^n\neq i \\  
  \left(f(x^{n-1})\right)_{S^n} & \text{ if }S^n=i.
\end{array}\right.
$$
\end{definition}

In other words, at the $n^{th}$ iteration, only the $S^{n}-$th cell is
\textquotedblleft  iterated\textquotedblright .  Note  that in  a more
general  formulation,  $S^n$  can   be  a  subset  of  components  and
$\left(f(x^{n-1})\right)_{S^{n}}$      can     be      replaced     by
$\left(f(x^{k})\right)_{S^{n}}$, where  $k<n$, describing for example,
delays  transmission~\cite{Robert1986}.  Finally,  let us  remark that
the term  ``chaotic'', in  the name of  these iterations,  has \emph{a
priori} no link with the mathematical theory of chaos, recalled below.

\subsubsection{Devaney's chaotic dynamical systems}
\label{subsection:Devaney}

Consider  a topological  space $(\mathcal{X},\tau)$  and  a continuous
function $f$ on $\mathcal{X}$.

\begin{definition}
  $f$ is said  to be \emph{topologically transitive} if,  for any pair
  of open sets $U,V \subset \mathcal{X}$, there exists $k>0$ such that
  $f^k(U) \cap V \neq \varnothing$.
\end{definition}

\begin{definition}
  An element  (a point) $x$  is a \emph{periodic element}  (point) for
  $f$  of period  $n\in \mathds{N}^*,$  if $f^{n}(x)=x$.
\end{definition}

\begin{definition}
  $f$ is said to be \emph{regular} on $(\mathcal{X}, \tau)$ if the set
  of periodic points for $f$  is dense in $\mathcal{X}$: for any point
  $x$ in $\mathcal{X}$, any neighborhood  of $x$ contains at least one
  periodic point.
\end{definition}

\begin{definition}
  $f$ is said  to be \emph{chaotic} on $(\mathcal{X},\tau)$  if $f$ is
  regular and topologically transitive.
\end{definition}

The   chaos   property  is   strongly   linked   to   the  notion   of
``sensitivity'', defined on a metric space $(\mathcal{X},d)$ by:

\begin{definition}
  \label{sensitivity} $f$ has \emph{sensitive dependence on initial conditions}
  if there  exists $\delta >0$  such that, for any  $x\in \mathcal{X}$
  and  any  neighborhood  $V$  of  $x$,  there  exists  $y\in  V$  and
  $n\geqslant 0$  such that $d\left(f^{n}(x),  f^{n}(y)\right) >\delta
  $.

  \noindent $\delta$ is called the \emph{constant of sensitivity} of $f$.
\end{definition}

Indeed, Banks  \emph{et al.}  have proven in~\cite{Banks92}  that when
$f$ is chaotic and $(\mathcal{X}, d)$  is a metric space, then $f$ has
the  property  of sensitive  dependence  on  initial conditions  (this
property was formerly  an element of the definition  of chaos). To sum
up, quoting Devaney in~\cite{Devaney}, a chaotic dynamical system ``is
unpredictable   because  of  the   sensitive  dependence   on  initial
conditions. It cannot be broken down or simplified into two subsystems
which do not interact because  of topological transitivity. And in the
midst  of this  random behavior,  we nevertheless  have an  element of
regularity''.  Fundamentally   different  behaviors  are  consequently
possible and occur in an unpredictable way.

\subsubsection{Chaotic iterations and Devaney's chaos}
\label{sec:topological}

In this section we give outline  proofs of the properties on which our
study of  chaotic neural networks  is based. The  complete theoretical
framework is detailed in~\cite{guyeux09}.

Denote by $\Delta $ the \emph{discrete boolean metric}, 
$\Delta(x,y)=0\Leftrightarrow x=y.$ Given a function 
$f:  \mathds{B}^{\mathsf{N}}\longrightarrow \mathds{B}^{\mathsf{N}}$, 
define the function $F_{f}:$ 
$\llbracket1;\mathsf{N}\rrbracket\times \mathds{B}^{\mathsf{N}}\longrightarrow 
\mathds{B}^{\mathsf{N}}$ 
such that
\begin{equation*}
F_{f}(k,E)=\left( E_{j}.\Delta (k,j)+f(E)_{k}.\overline{\Delta (k,j)}\right)_{j\in \llbracket1;\mathsf{N}\rrbracket},
\end{equation*}

\noindent  where  +  and  .   are the  boolean  addition  and  product
operations,   $\overline{x}$  is   for  the   negation  of   $x$. 

Consider                the                phase                space
$\mathcal{X}=\llbracket1;\mathsf{N}\rrbracket^{\mathds{N}}\times
\mathds{B}^{\mathsf{N}}$ and the map
\begin{equation*}
G_{f}\left( S,E\right) =\left( \sigma (S),F_{f}(i(S),E)\right).
\end{equation*}
\noindent where the
\emph{shift}    function   is    defined    by   $\sigma:(S^{n})_{n\in
\mathds{N}}\in   \mathbb{S}  \mapsto   (S^{n+1})_{n\in  \mathds{N}}\in
\mathbb{S}$,  and the  \emph{initial function}  $i$ is  the  map which
associates   to   a  sequence,   its   first  term:   $i:(S^{n})_{n\in
\mathds{N}}\in        \mathbb{S}        \mapsto       S^{0}        \in
\llbracket1;\mathsf{N}\rrbracket$.

Thus chaotic iterations can be described by the following iterations\cite{guyeux09}
\begin{equation*}
\left\{
\begin{array}{l}
X^{0}\in \mathcal{X} \\
X^{k+1}=G_{f}(X^{k}).
\end{array}
\right.
\end{equation*}

Let us define a new distance between two points 
$(S,E),(\check{S},\check{E} )\in \mathcal{X}$ by
\begin{equation*}
d((S,E);(\check{S},\check{E}))=d_{e}(E,\check{E})+d_{s}(S,\check{S}),
\end{equation*}

where
\begin{itemize}
\item
$\displaystyle{d_{e}(E,\check{E})}=\displaystyle{\sum_{k=1}^{\mathsf{N}}\Delta
(E_{k},\check{E}_{k})} \in \llbracket 0 ; \mathsf{N} \rrbracket$
\item
$\displaystyle{d_{s}(S,\check{S})}=\displaystyle{\dfrac{9}{\mathsf{N}}\sum_{k=1}^{\infty
}\dfrac{|S^{k}-\check{S}^{k}|}{10^{k}}} \in [0 ; 1].$
\end{itemize}

This new  distance has been  introduced in \cite{guyeux09}  to satisfy
the following requirements. When the number of different cells between
two systems is increasing, then their distance should increase too. In
addition, if two  systems present the same cells  and their respective
strategies start with the same  terms, then the distance between these
two points must be small because the evolution of the two systems will
be the same  for a while.  The distance  presented above follows these
recommendations. Indeed,  if the floor value  $\lfloor d(X,Y)\rfloor $
is  equal  to $n$,  then  the systems  $E,  \check{E}$  differ in  $n$
cells. In addition, $d(X,Y) - \lfloor d(X,Y) \rfloor $ is a measure of
the  differences   between  strategies  $S$   and  $\check{S}$.   More
precisely, this  floating part is less  than $10^{-k}$ if  and only if
the first $k$ terms of the  two strategies are equal. Moreover, if the
$k^{th}$  digit  is  nonzero,  then  the $k^{th}$  terms  of  the  two
strategies are different.

\medskip

It  is proven in  \cite{guyeux09} by  using the  sequential continuity
that                the               vectorial               negation
$f_{0}(x_{1},\hdots,x_{\mathsf{N}})=(\overline{x_{1}},\hdots,
\overline{x_{\mathsf{N}}})$ satisfies the following proposition:

\begin{proposition}
\label{continuite} $G_{f_0}$ is a continuous function on $(\mathcal{X},d)$.
\end{proposition}

It is then checked, also  in \cite{guyeux09}, that in the metric space
$(\mathcal{X},d)$, the vectorial negation fulfill the three conditions
for Devaney's  chaos: regularity, transitivity,  and sensitivity. This
has led to the following result.

\begin{proposition}
$G_{f_0}$ is a chaotic map on $(\mathcal{X},d)$ in the sense of Devaney.
\end{proposition}

\subsection{Neural Networks}
\label{sec:NN}

\begin{figure}[t]
\centering
\includegraphics[width=2.5in]{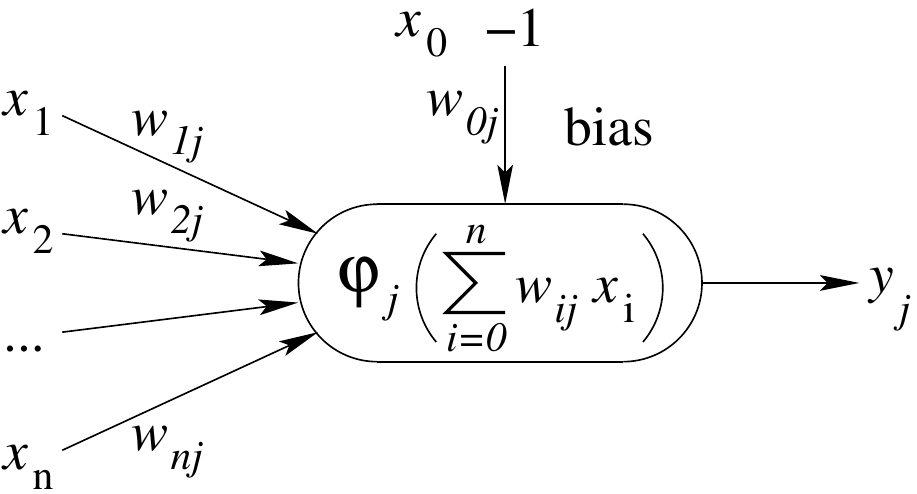}
\caption{Description of a neuron}
\label{neuron}
\end{figure}

An artificial  neural network is  a set of simple  processing elements
called  neurons   that  are  interconnected,  usually   with  a  layer
structure.  It takes some input  values and produces some output ones.
Like  a biological  neural  network, the  connections between  neurons
influence the  outputs given by  the artificial network.  Thanks  to a
training  process,  an ANN  is  able  to  learn complex  relationships
between  inputs   and  outputs.   A  neuron~$j$   computes  an  output
$y=\varphi(x,w)$ where $\varphi()$ is  the activation function, $x$ is
the input  vector, and $w$  the parameter vector.   $w$ can be  used to
parameterize  $\varphi$ or  the neuron  inputs. In  this last  case it
means that the connections are  weighted and a vector $w$ component is
then referred to as  a synaptic weight.  Figure~\ref{neuron}~describes
a neuron~$j$ with weighted connections. Its output $y_j$ satisfies:
\begin{equation}
y_j = \varphi_j\left(\sum_{i=1}^n w_{ij} x_i+b_j\right)
    = \varphi_j\left(\sum_{i=0}^n w_{ij} x_i\right)
\end{equation}
where  $x_0=-1$,   $x=\left(x_1,\dots,x_n\right)$,  and  $w_{0j}=-b_j$
defines the bias value.

Neural networks  have a layered  architecture, but they may  differ in
the way the  output of a neuron affect himself. In  fact, based on the
connection graph,  two kinds of  networks can be  distinguished: those
having at least  one loop and those without any  one. A neural network
which exhibits  a loop  is called a  feedback (or  recurrent) network,
whereas a network belonging to  the second class is said feed-forward.
Obviously, a  feedback network can be  seen as a  dynamical system. In
the following sections, we use  a recurrent version of the multi-layer
perceptron  (MLP),  a  well-known   ANN  architecture  for  which  the
universal approximation  property has been proven  in the feed-forward
context   \cite{DBLP:journals/jpdc/Cybenko89}.    Typically,   a   MLP
consists in a layer of input  neurons, in one or more layers of hidden
neurons, and a layer of output neurons. Since an input neuron is simply
used as  a channel to  dispatch an input  to each neuron of  the first
hidden layer, we will not  further consider the input layer.  Usually,
the neurons  of a given layer have  similar characteristics and  each one is
fully connected to the next layer. Finally, it can be noticed that the
number of  inputs and  output neurons is  completely specified  by the
considered  problem,  while  the  number  of  hidden  neurons  depends
directly on the  complexity of the relationships to  be learned by the
ANN.

As   said  previously,  a   neural  network   is  designed   to  model
relationships between inputs  and outputs.  In order to  find a proper
modeling, an ANN  must be trained so that it  provides the desired set
of output vectors. The  training (or learning) process consists mainly
in  feeding the  network  with  some input  vectors  and updating  the
neurons parameters (weights and bias  value) using a learning rule and
some information  which reflects the quality of  the current modeling.
When the  expected output  vectors ($D_k$) are  known in  advance, the
quality   can   be    expressed   through   the   Mean-Squared   Error
\cite{lecun-98b}:
\begin{equation}
\mbox{MSE} = \frac{1}{2 N} \sum_{k=1}^N \left(D_k-Y_k\right)
\end{equation}
where $N$ is the number of input-output vector pairs used to train the
ANN (the  pair set is called  the training or learning  set) and $Y_k$
denotes  an output vector  produced by  the output  layer for  a given
input vector $X_k$.  Consequently,  in that case the training process,
which  is  said  supervised,  results  in  an  optimization  algorithm
targeted  to  find the  weights  and  biases  that minimize  the  MSE.
Various  optimization  techniques  exist,  they have  given  raise  to
distinct training  algorithms performing iterative  parameters update.
Gradient   based  methods   are  particularly   popular  due   to  the
backpropagation  algorithm, but  they are  sensitive to  local minima.
Heuristics like  simulated annealing or  differential evolution permit
to  find a  global  minimum, but  they  have a  slow convergence.   To
control the training process, two  methods are the most commonly used:
firstly the number of iterations, also called epochs, reaches an upper
bound, secondly the MSE goes below a threshold value.

\section{Related Work}
\label{sec:RelatedW}

Since a  while neuroscientists discuss  the existence of chaos  in the
brain. In the context of artificial neural networks, this interest has
given  raise  to various  works  studying  the  modeling of  chaos  in
neurons.  The  chaotic neuron  model designed by  Aihara {\em  et al.}
\cite{Aihara1990333}  is  particularly used  to  build chaotic  neural
networks.  For example, in \cite{DBLP:conf/esann/CrookS01} is proposed
a feedback ANN  architecture which consists of two  layers (apart from
the input  layer) with  one of them  composed of chaotic  neurons.  In
their experiments, the authors  showed that without any input sequence
the activation  of each chaotic  neuron results in a  positive average
Lyapunov exponent, which means a true chaotic behavior.  When an input
sequence is given iteratively to the network the chaotic neurons reach
stabilized   periodic  orbits   with  different   periods,   and  thus
potentially provide a recognition  state.  Similarly, the same authors
have  recently  introduced  another   model  of  chaotic  neuron:  the
non-linear dynamic state  (NDS) neuron, and used it  to build a neural
network which is able  to recognize learned stabilized periodic orbits
identifying patterns \cite{Crook2007267}.

Today, another field of research in which chaotic neural networks have
received  a lot  of  attention  is data  security.   In fact,  chaotic
cryptosystems are  an appealing alternative  to classical ones  due to
properties such  as sensitivity  to initial conditions  or topological
transitivity.   Thus  chaotic  ANNs  have  been  considered  to  build
ciphering  methods,  hash  functions,  digital  watermarking  schemes,
pseudo-random  number generators,  etc.  In  \cite{dalkiran10}  such a
cipher   scheme  based   on  the   dynamics  of   Chua's   circuit  is
proposed. More precisely, a feed-forward MLP with two hidden layers is
built to  learn about 1500~input-output vector pairs,  where each pair
is obtained  from the three nonlinear  ordinary differential equations
modeling the circuit.  Hence, the proposed chaotic neural network is a
network which is trained to  learn a true chaotic physical system.  In
the cipher scheme the ANN plays the role of chaos generator with which
the  plain-text will be  merged. Untrained  neural networks  have been
also considered to define  block ciphering \cite{Lian20091296} or hash
functions      \cite{springerlink:10.1007/s00521-010-0432-2}.      The
background  idea is  to exploit  the inherent  properties of  the ANNs
architecture such as diffusion and confusion.
 
\section{A First Recurrent Neural Network \\ Chaotic According to Devaney}
\label{sec:KoANN}

\subsection{Defining a First Chaotic Recurrent Neural Network}
\label{sec:Formalizing}

We  will now  explain  how to  build  a chaotic  neural network  using
chaotic iterations.

Let  us   reconsider  the  vectorial  negation   function  denoted  by
$f_0:\mathds{B}^\mathsf{N}   \to    \mathds{B}^\mathsf{N}$   and   its
associated  map $F_{f_0}:\llbracket  1;  \mathsf{N} \rrbracket  \times
\mathds{B}^\mathsf{N}  \to  \mathds{B}^\mathsf{N}$.   Firstly,  it  is
possible to  define a MLP  which recognize $F_{f_0}$. That  means, for
all   $(k,x)  \in   \llbracket  1   ;  \mathsf{N}   \rrbracket  \times
\mathds{B}^\mathsf{N}$, the response of  the output layer to the input
$(k,x)$  is  $F_{f_0}(k,x)$.   Secondly,   the  output  layer  can  be
connected   to    the   input   layer    as   it   is    depicted   in
Figure~\ref{perceptron}, leading to  a global recurrent neural network
working as follows:

\begin{figure}[!t]
\centering
\includegraphics[width=3.5in]{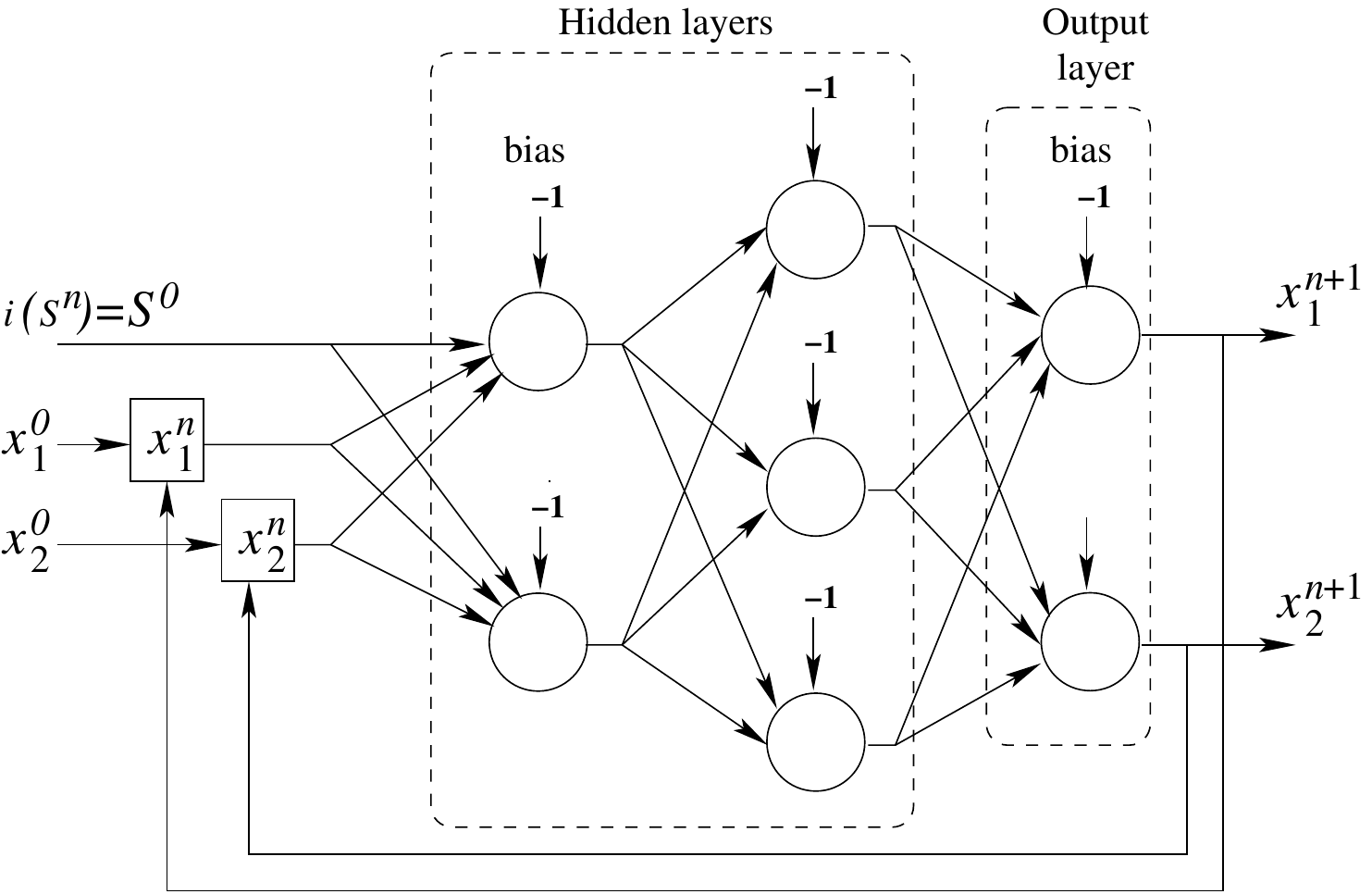}
\caption{Example of global recurrent neural network modeling function $F_{f_0}$ such that 
  $x^{n+1}=\left(x^{n+1}_1,x^{n+1}_2\right)=F_{f_0}\left(i(S^n),\left(x^n_1,x^n_2\right)\right)$}
\label{perceptron}
\hfil
\end{figure}

\begin{itemize}
\item At the  initialization stage, the ANN receives  a boolean vector
$x^0\in\mathds{B}^\mathsf{N}$ as input  state, and $S^0 \in \llbracket
1;\mathsf{N}\rrbracket$  in its  input integer  channel  $i()$.  Thus,
$x^1 =  F_{f_0}(S^0, x^0)\in\mathds{B}^\mathsf{N}$ is  computed by the
neural network.
\item This state $x^1$ is  published as an output. Additionally, $x^1$
is sent back to  the input layer, to act as boolean  state in the next
iteration.
\item At  iteration number $n$, the recurrent  neural network receives
the  state  $x^n\in\mathds{B}^\mathsf{N}$ from  its  output layer  and
$i\left(S^n\right)  \in  \llbracket  1;\mathsf{N}\rrbracket$ from  its
input  integer  channel  $i()$.   It  can thus  calculate  $x^{n+1}  =
F_{f_0}(i\left(S^n\right),  x^n)\in\mathds{B}^\mathsf{N}$,  which will
be the new output of the network.
\end{itemize}

In  this way,  if the  initial state  $x^0\in\mathds{B}^\mathsf{N}$ is
sent   to   the   network   with   a  sequence   $S   \in   \llbracket
1;\mathsf{N}\rrbracket^\mathds{N}$   applied  in  the   input  integer
channel   $i()$,   then    the   sequence   $\left(x^n\right)_{n   \in
\mathds{N}^*}$ of  the outputs is  exactly the same than  the sequence
obtained    from   the    following   chaotic    iterations:   $x^0\in
\mathds{B}^{\mathsf{N}}$ and
$$     
\forall n \in \mathds{N}^{\ast}, \forall i \in \llbracket1;\mathsf{N}\rrbracket,
x_i^n=\left\{
\begin{array}{ll}
  x_i^{n-1} &  \text{ if  }S^n\neq i \\  
  \left(f_0(x^{n-1})\right)_{S^n} & \text{ if }S^n=i.
\end{array}\right.
$$

From a mathematical viewpoint, the  MLP defined in this subsection and
chaotic  iterations  recalled  above   have  the  same  behavior.   In
particular,  given  the same  input  vector  $\left(x^0, (S^n)_{n  \in
\mathds{N}}\right)$,   they    produce   the   same    output   vector
$\left(x^n\right)_{n  \in  \mathds{N}^*}$:  they  are  two  equivalent
reformulations of  the iterations of $G_{f_0}$ in  $\mathcal{X}$. As a
consequence, the behavior of  our MLP faithfully reflects the behavior
of $G_{f_0}$ which is chaotic according to Devaney.

\subsection{Improving the Variety of Chaotic Recurrent Neural Networks}

The approach  proposed to build chaotic neural  networks, explained in
the  previous  subsection, is  not  restricted  to  an adhoc  function
$f_0:\mathds{B}^\mathsf{N}  \to   \mathds{B}^\mathsf{N}$,  it  can  be
generalized  as  follows. The  function  $F_{f_0}$  associated to  the
vectorial  negation $f_0$,  which has  been recognized  by  the neural
network,  can  be replaced  by  any  functions  $F_f: \llbracket  1  ;
\mathsf{N}     \rrbracket     \times     \mathds{B}^\mathsf{N}     \to
\mathds{B}^\mathsf{N}$  such  that the  chaotic  iterations $G_f$  are
chaotic, as defined by Devaney.

To be able to define functions  that can be used in this situation, we
must firstly introduce the graph of iterations of a given function $f:
\mathds{B}^\mathsf{N}  \rightarrow  \mathds{B}^\mathsf{N},  x  \mapsto
(f_1(x), \ldots, f_n(x))$.

Let be  given a configuration  $x$. In what follows  the configuration
$N(i,x)  =   (x_1,\ldots,\overline{x_i},\ldots,x_n)$  is  obtained  by
switching the  $i-$th component of $x$. Intuitively,  $x$ and $N(i,x)$
are neighbors.   The chaotic  iterations of the  function $f$  can be
represented by the graph $\Gamma(f)$ defined below.

\begin{definition}[Graph of iterations]
In the  oriented \emph{graph of iterations}  $\Gamma(f)$, vertices are
configurations of $\mathds{B}^\mathsf{N}$ and  there is an arc labeled
$i$ from $x$ to $N(i,x)$ iff $F_f(i,x)$ is $N(i,x)$.
\end{definition}
We have proven in \cite{GuyeuxThese10} that:
\begin{theorem}
\label{Th:Caracterisation des IC chaotiques} Functions $f :
\mathds{B}^{n}  \to   \mathds{B}^{n}$  such  that   $G_f$  is  chaotic
according to Devaney, are functions such that the graph $\Gamma(f)$ is
strongly connected.
\end{theorem}

Since it is easy to check whether a graph is strongly connected, we can
use  this   theorem  to  discover   new  functions  \linebreak   $f  :
\mathds{B}^\mathsf{N} \rightarrow \mathds{B}^\mathsf{N}$ such that the
neural network associated to  $G_f$ behaves chaotically, as defined by
Devaney.

\subsection{The Discovery of New Chaotic Neural Networks}

Considering Theorem~\ref{Th:Caracterisation des  IC chaotiques}, it is
easy                   to                  check                  that
$f_{1}(x_{1},\hdots,x_{\mathsf{N}})=(\overline{x_{1}},x_1,         x_2,
\hdots, x_{\mathsf{N}-1})$ is such that $G_{f_1}$ behaves chaotically,
as defined  by Devaney.  Consequently, we can  now obtain  two chaotic
neural networks by learning either $F_{f_0}$ or $F_{f_1}$.
 
To support  our approach, a  set of illustrative examples  composed of
five  neural   networks  is  given.  The  three   first  networks  are
respectively defined by:
\begin{itemize}
\item $f_{0,1}(x_{1},x_2,x_3,x_4)=(\overline{x_{1}},\overline{x_{2}},\overline{x_{3}},\overline{x_{4}})$,
\item $f_{0,2}(x_{1},x_2,x_3)=(\overline{x_{1}},\overline{x_{2}},\overline{x_{3}})$,
\item $f_{1,1}(x_{1},x_2,x_3)=(\overline{x_{1}},x_{1},x_{2})$,
\end{itemize}
while the last ones are defined by:
\begin{itemize}
\item $g_{0}(x_{1},x_2,x_3)=(x_{1},x_{2},x_{3})$,
\item $g_{1}(x_{1},x_2,x_3)=(\overline{x_{1}},x_{2},x_{3})$.
\end{itemize}
Due to  Theorem \ref{Th:Caracterisation  des IC chaotiques},  the ANNs
associated to  $f_{0,1}$, $f_{0,2}$ and  $f_{1,1}$ behave chaotically,
as defined  by Devaney. Whereas  it is not  the case for  the networks
based   on  the   boolean   functions  $g_{0}$   and  $g_{1}$,   since
$\Gamma(g_0)$ and $\Gamma(g_1)$ are not strongly connected.

\subsection{Experimental results}

Among the  five neural networks  evoked in the previous  subsection we
decided to study the training process of three of them. Note also that
for each neural network we  have considered MLP architectures with one
and two  hidden layers,  with in the  first case different  numbers of
hidden neurons  (sigmoidal activation).   Thus we will  have different
versions  of a neural  network modeling  the same  iteration function.
Only the size and number of hidden layer may change, since the numbers
of inputs  and output neurons (linear activation)  are fully specified
by  the   function.   The  neural  networks  are   trained  using  the
quasi-Newton         \linebreak         L-BFGS         (Limited-memory
Broyden-Fletcher-Goldfarb-Shanno)  algorithm in  combination  with the
Wolfe  linear search.  The  training is  performed until  the learning
error (MSE) is lower than a chosen threshold value ($10^{-2}$).

\begin{table}[!t]
\renewcommand{\arraystretch}{1.3}
\caption{Outline  of  the  results  from several  iteration  functions
learning using different recurrent MLP architectures}
\label{results}
\centering
\begin{tabular}{|c||c|c||c|c|}
\hline 
  & \multicolumn{4}{c|}{One hidden layer} \\
\cline{2-5}
  & \multicolumn{2}{c||}{8 neurons} & \multicolumn{2}{|c|}{10 neurons} \\
\hline
Function & Mean epoch & Success rate & Mean epoch & Success rate \\
\hline
$f_{0,2}$ & 82.21 & 100\% & 73.44 & 100\% \\
$f_{1,1}$ & 76.88 & 100\% & 59.84 & 100\% \\
$g_1$ & 36.24 & 100\% & 37.04 & 100\% \\
\hline
\hline
  & \multicolumn{4}{c|}{Two hidden layers: 8 and 4 neurons} \\
\cline{2-5}
  & \multicolumn{2}{c|}{Mean epoch number} & \multicolumn{2}{|c|}{Success rate} \\
\hline
$f_{0,2}$ & \multicolumn{2}{c|}{203.68} & \multicolumn{2}{c|}{76\%} \\
$f_{1,1}$ & \multicolumn{2}{c|}{135.54} & \multicolumn{2}{c|}{96\%} \\
$g_1$ & \multicolumn{2}{c|}{76.56} & \multicolumn{2}{c|}{100\%} \\
\hline
\end{tabular}
\end{table}

Table~\ref{results} gives for each  considered neural network the mean
number  of  epochs needed  to  train them  and  a  success rate  which
reflects a successful  training in less than 1000  epochs. Both values
are computed  considering 25 trainings with random  weights and biases
initialization.   These results  highlight  several points.   Firstly,
various MLP  architectures can learn  a same iteration  function, with
obviously a best suited one  (a hidden layer composed of ten sigmoidal
neurons).  In  particular the two  hidden layer structure seems  to be
too  complex  for the  functions  to  be  learned. Secondly,  training
networks so that  they behave chaotically seems to  be more difficult,
since   they   need  in   average   more   epochs   to  be   correctly
trained.  However, the  relevance of  this point  needs to  be further
investigated.  Similarly, there  may be  a link  between  the training
difficulty  and  the  disorder   (evaluation  of  their  constants  of
sensitivity,  expansivity,  etc.)   induced  by  a  chaotic  iteration
function.

\section{Conclusion and future work}
\label{sec:Conclusion}

Many chaotic neural networks  have been developed for different fields
of application, in particular for data security purpose where they are
used  to   define  ciphering  methods,  hash  functions   and  so  on.
Unfortunately, the proposed networks are usually claimed to be chaotic
without  any  proof.  In  this  paper  we  have presented  a  rigorous
mathematical  framework  which   allows  us  to  construct  artificial
networks  proven chaotic,  according  to Devaney.   More precisely,  a
correspondence between chaotic iterations, which are a particular case
of topological chaos in sense of Devenay, and MLP neural networks with
a  global feedback is  established.  In  fact, we  have shown  that an
iteration function  is chaotic if  its graph of iteration  is strongly
connected (a property easily checked), and that a global recurrent MLP
can learn such  a function. Future research will  study more carefully
the performance of the training process and alternative neural network
architectures.





\bibliographystyle{IEEEtran}
\bibliography{article,mabase}
%

\end{document}